\documentclass[conference]{IEEEtran}
\IEEEoverridecommandlockouts

\usepackage[top=0.6in, bottom=.8in, left=0.62in, right=0.62in]{geometry}
\usepackage{setspace} 
\usepackage{cite}
\usepackage{amsmath,amssymb,amsfonts}
\usepackage{graphicx}
\usepackage{textcomp}
\usepackage{xcolor}

\usepackage{subcaption}
\usepackage[font={small,it}]{caption}
\usepackage{float}

\usepackage{subcaption}
\usepackage[font={small,it}]{caption}
\usepackage{float}
\usepackage{booktabs}
\usepackage{algorithm}
\usepackage[noend]{algpseudocode}

\usepackage{latexsym}
\usepackage{caption}
\usepackage{mathptmx}
\usepackage{listings}
\usepackage{float}
\usepackage{wrapfig}
\usepackage{multirow}
\usepackage{makecell}

\usepackage{tikz}
\usepackage{verbatim}
\usepackage{scalefnt}
\usepackage{adjustbox}

\def\BibTeX{{\rm B\kern-.05em{\sc i\kern-.025em b}\kern-.08em
    T\kern-.1667em\lower.7ex\hbox{E}\kern-.125emX}}

\begin{document}

\title{
Semi-supervised 3D Video Information Retrieval with Deep Neural Network and Bi-directional Dynamic-time Warping Algorithm
}

\author{
\IEEEauthorblockN{1\textsuperscript{st} Yintai Ma}
\IEEEauthorblockA{\textit{Dept. of Industrial Enginering and Management Science} \\
\textit{Northwestern University}\\
Evanston, United States \\
yintaima2022@u.northwestern.edu}
\and
\IEEEauthorblockN{2\textsuperscript{nd} Diego Klabjan}
\IEEEauthorblockA{\textit{Dept. of Industrial Enginering and Management Science} \\
\textit{Northwestern University}\\
Evanston, United States \\
d-klabjan@northwestern.edu}
}

\maketitle

\begin{abstract}
This paper presents a novel semi-supervised deep learning algorithm for retrieving similar 2D and 3D videos based on visual content. The proposed approach combines the power of deep convolutional and recurrent neural networks with dynamic time warping as a similarity measure. The proposed algorithm is designed to handle large video datasets and retrieve the most related videos to a given inquiry video clip based on its graphical frames and contents. We split both the candidate and the inquiry videos into a sequence of clips and convert each clip to a representation vector using an autoencoder-backed deep neural network. We then calculate a similarity measure between the sequences of embedding vectors using a bi-directional dynamic time-warping method. This approach is tested on multiple public datasets, including CC\_WEB\_VIDEO, Youtube-8m, S3DIS, and Synthia, and showed good results compared to state-of-the-art. The algorithm effectively solves video retrieval tasks and outperforms the benchmarked state-of-the-art deep learning model.
\end{abstract}


\begin{figure}[ht]
\begin{center}
\centerline{\includegraphics[width=\columnwidth]{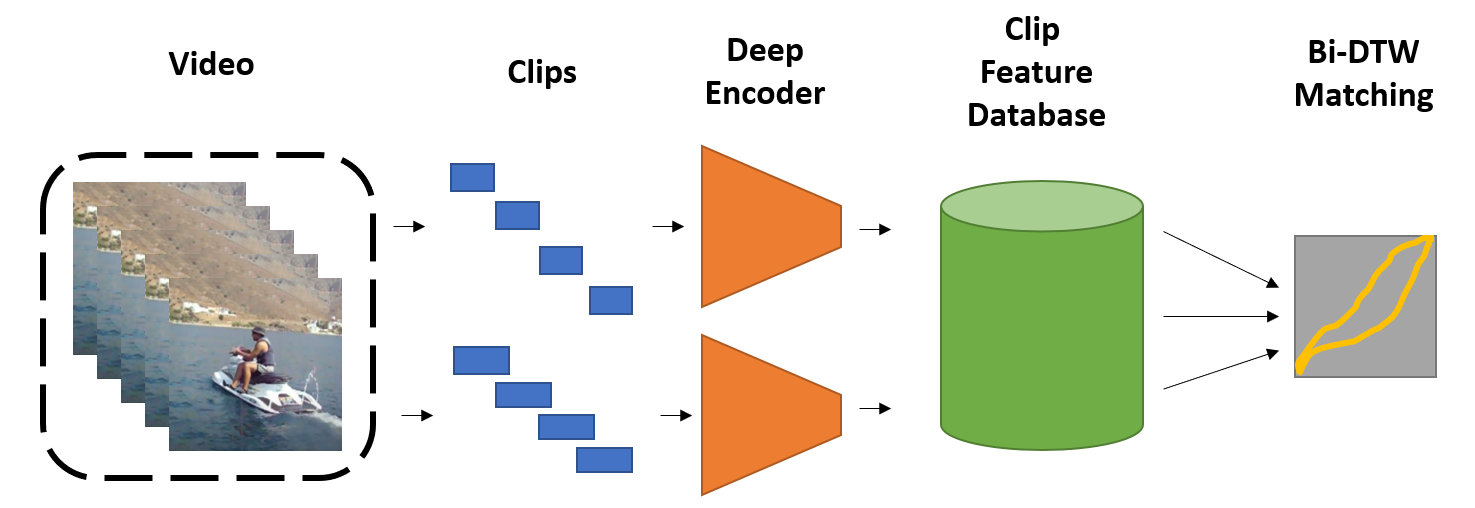}}
\caption{The proposed video representation learning pipeline first splits videos into short clips at a fixed number of frames. Then the proposed method converts clips to embedded feature vectors by a deep neural network encoder. Finally, we store all feature vectors in a database and use a bi-directional dynamic time-warping method to retrieve a list of candidates.}
\label{fig-video-representation-learning}
\end{center}
\vskip -0.2in
\end{figure}

\section{Introduction}
\label{introduction}

In recent years, the exponential growth in online video data has made efficient retrieval of visually similar videos increasingly challenging. To address this, we propose a novel semi-supervised deep learning framework for retrieving similar 2D and 3D videos.

Our video retrieval approach represents videos as sequences of embedding vectors generated by a deep neural network encoder. It begins by splitting both candidate and query videos into fixed-length consecutive clips. Each clip is fed into the encoder to produce a representation vector. The sequence of embedding vectors for the full video is then compared to candidate videos using a bidirectional dynamic time warping similarity measure. By transforming videos into informative embedding sequences, and leveraging deep neural networks with dynamic time warping, our framework can effectively retrieve visually similar 2D and 3D videos.

A video embedding is created through the use of deep convolutional and recurrent neural networks, which are designed to extract rich and discriminative features from video clips. To optimize the video retrieval performance, we adopt a two-stage training approach. First, we pre-train the model unsupervised on a larger and relevant video dataset. Second, we fine-tune the model with a triplet loss function in a supervised manner, further enhancing its ability to perform video retrieval.

Finally, we compute the similarity between the sequences of embedding vectors between the query video and candidate videos using a variant of the dynamic time warping method. The bi-directional Dynamic Time Warping (Bi-DTW) method has been employed as a means to address the limitations of the standard Dynamic Time Warping (DTW) in video embedding matching. The standard DTW algorithm, although effective at time series alignment, operates only in a single, forward direction. This unidirectional characteristic can potentially limit the accuracy of its matching results, particularly in applications such as video retrieval where the temporal structure of the data is complex and multidimensional. To mitigate these limitations, Bi-DTW was developed, with a distinguishing feature being its capacity to facilitate matching from both forward and backward directions. The rationale for this approach is rooted in the intuition that different portions of a video may align more effectively when approached from various temporal perspectives. Therefore, by integrating both forward and backward alignments, Bi-DTW improves the accuracy of video retrieval by ensuring a more robust, comprehensive temporal match. It presents a significant improvement over the conventional DTW, especially in complex, time-structured applications like video retrieval where the objective is to maximize the accuracy of the matching results. This allows us to determine the degree of similarity between the two videos, and effectively rank the candidate videos based on their relevance to the query video.

We evaluate the proposed method on several publicly available datasets, including CC\_WEB\_VIDEO, Youtube-8m, S3DIS, and Synthia, and show good results in comparison to state-of-the-art.

\subsection{Contributions}

We present a cutting-edge approach to video information retrieval by introducing a bi-directional dynamic time-warping method for determining the similarity between video inquiries. This innovative technique effectively tackles the temporal dimension of videos, resulting in improved accuracy and efficiency of the retrieval process.

Furthermore, this research encompasses the handling of 3D video inquiry, introducing both a novel 3D network architecture that expands upon 2D video information retrieval models and a method to incorporate 3D video data as an additional depth layer. This comprehensive framework offers a more robust solution for retrieving 3D video data.

We also introduce a sample retraining method that effectively addresses the challenge of handling difficult video pairs during the training phase. This method increases the number of previously under-studied data points, ultimately leading to improved performance and accuracy.

The proposed method in this paper demonstrates a remarkable advancement in video information retrieval, offering the potential to significantly enhance video search and retrieval systems across a wide range of applications. Our findings make video search more accessible and valuable to a broad range of users.

\section{Related Works}

\label{sec:v2vir:lr}

The field of video information retrieval has been the subject of extensive research in recent years, due to the growing demand for effective and efficient methods for searching and retrieving video content. A wide range of techniques and approaches have been proposed to address the challenges associated with video retrieval, including traditional methods based on feature extraction and machine learning, as well as more recent deep learning-based methods. The literature in this area is vast, encompassing a wide range of topics, including the representation and encoding of video data, the development of effective similarity measures, and the handling of temporal and 3D information in video data. In this literature review, we provide an overview of the current state-of-the-art in video information retrieval, highlighting key contributions and developments in the field, and pointing to similarities and differences with our work.

The foundation of seq2seq models is simple neural network encoder and decoder recurrent models, such as LSTMs \cite{Srivastava2015} and GRUs \cite{Cho2014}. Studies on the hierarchy for these encoders show that the better encoder networks in the model should lead to better results \cite{Zhang}. The proposed model builds upon this foundation, further advancing the encoder networks to enhance the overall performance. Consequently, in this work, we  improve the state-of-the-art in this area by developing a deep convolutional and recurrent model using recent developments in the related areas. 

\textbf{Transformer} architecture has been increasingly popular in recent years. It was applied to understand video data and perform video retrieval. The transformer, initially proposed for language understanding tasks, is a powerful architecture capable of capturing long-range dependencies and handling sequential data. One of the key features of the transformer is the self-attention mechanism, which allows the model to weigh the importance of different parts of the input data when making predictions. Several studies have applied transformer-based architectures to video understanding tasks such as action recognition, video captioning, and video retrieval. For example, Girdhar proposes a transformer-based architecture for video action recognition \cite{girdhar_video_2019}. Similarly, Im and Choi propose a transformer-based model for video captioning \cite{im_uat_2022}. The transformer has proven to be particularly useful in video understanding tasks, where the model needs to understand the relationships between different frames in a video. Our proposed method builds on the strengths of the transformer and further improves upon it by utilizing bi-directional dynamic time-warping to enhance the accuracy of video retrieval.

\textbf{Video Retrieval.} There is a variety of retrieval tasks and definitions in the multimedia community concerning the video retrieval problem. These vary with respect to the degree of similarity that determines whether a pair of videos are considered related and range from Near-Duplicate Video Retrieval (NDVR) with a very narrow scope where only almost identical videos are considered positive pairs \cite{Wu2007}, to very broad where videos from the same event \cite{Revaud2013} in Event Video Retrieval (EVR) or with the same semantics \cite{Basharat2008} are labeled as related. In the copy detection problem, given a query video, only videos containing nearly identical copies of it should be retrieved. Similar videos from the same incident should be considered irrelevant in such a scenario.
On the other hand, problems such as news-oriented retrieval have radically different needs. Deep learning methods have been applied to certain applications such as face video retrieval \cite{choi_face_2021} or as a general method \cite{kordopatis-zilos_near-duplicate_2017}. As an extension to the 3D case, Deng has proposed a pipeline to handle the 3D video data for retrieval purposes \cite{deng_3d-csl_2022}. However, there does not seem to be a strong consensus among researchers about the cases where the videos are all unlabeled, and the task is to retrieve videos sharing scenarios and semantic meanings.  While there are a variety of retrieval tasks and definitions in the multimedia community, the proposed approach is a unified framework that can handle both 2D and 3D cases, therefore we  focus on maximizing accuracy in video retrieval tasks where the temporal structure of the data is complex and multidimensional.

\textbf{Convolutional Neural Networks (CNN's).} CNNs have been successfully applied to many tasks related to similarity video search \cite{Karpathy2014,Zou2012,Bazzani2011,Mobahi2009}. Unlike Deep Neural Networks, CNNs effectively exploit the structural locality in the spectral feature space. CNNs use filters with shared weights and pooling to give the model better spectral and temporal invariance properties. It typically generates more stable features compared to DNNs. The recent deep CNNs also show their superiority in image-related tasks compared to previous CNNs. Deep CNNs have been used for many tasks relating to video, such as predicting the next frame \cite{Palm2012}, learning invariant features from video \cite{Chen2010, Srivastava2015} or video classification \cite{Karpathy2014}. Many training and modeling tricks, such as Residual Network \cite{Kim}, have been developed to enable training for such deep networks. While CNN's have been applied to many tasks related to similarity video search, the proposed model takes advantage of bi-directional dynamic time-warping and recurrent neural network to better capture the temproal nature of the videos and improve the retrieval process.

\textbf{Traditional Image Retrieval} has been extensively explored as query by example or near-duplicate detection with high potential for the medical community \cite{rui_image_1999,Datta2008}. 
In literature, many research publications in image retrieval are based on non-deep learning methods.
A competition for image-based retrieval was organized between 2004 and 2013. This case differs from the one addressed in this work because they were defined with 1-7 sample images accompanied by text. In the 2013 edition \cite{garcia_seco_de_herrera_overview_2014}, the best textual run by Herrera \cite{herrera_comparing_2015} achieved the same performance as the best technique using both textual and visual features. As in previous years, visual-only approaches achieved much lower performance than textual and multimodal techniques. The best visual-based solution is based on the Color and Edge Directivity Descriptor (CEDD), a fuzzy color and texture histogram, and a Color Layout Descriptor \cite{ozturkmenoglu_effects_2013}. Content-based image retrieval in the medical domain has been addressed from low-level wavelet-based visual signatures \cite{quellec_wavelet_2010} to high-level concept detectors \cite{Rahman2011}. However, these methods are designed explicitly with specific domain knowledge. The proposed model extends these techniques to the video retrieval problem and further enhances them with bi-directional dynamic time-warping.

\textbf{Auto-encoder} has a long history \cite{Ballard1987} of pre-training artificial neural networks and is widely used in recent models \cite{Bengio2009}. Although this concept is rarely used by other deep learning models in this area, we found it to be fundamentally important for our semi-supervised learning purpose. In the proposed approach, we leverage this concept for semi-supervised learning in video retrieval tasks.

\textbf{Representation Learning.} Much of computer vision is about learning the representation, such as learning high-level image classification \cite{Russakovsky2015}, object relationships  \cite{Meister2018}, or point-wise correspondences  \cite{Bansal2018, kanazawa_warpnet_2016, Liu2011}. However, there has been relatively little work on learning representation for aligning the content of different videos. In this work, we identify similar videos in the dataset, which is essentially aligning the content of two videos in a self-supervised manner, and do the automatic alignment of the visual data without any additional supervision. Although much of computer vision involves learning representation, we extend this notion to align the content of different videos in a self-supervised manner, thanks to the proposed bi-directional dynamic time-warping method.

\textbf{ConvLSTM.} Shi introduces convolutional LSTM (ConvLSTM) as an extension to the original LSTM, where the inner product is replaced by convolution operation in both input-to-state and state-to-state transitions \cite{Shi}. ConvLSTM effectively maintains the structural representations in the output and cell state. It has been shown to be a better tool than a fully connected LSTM layer for maintaining structural locality and more prone to overfitting. Besides, it reduces the number of parameters within the layer and enables potential more computations for better generalization. The ConvLSTM has been a substantial tool in maintaining structural locality and preventing overfitting. However, in the research, we utilize the bi-directional dynamic time-warping approach, offering a more comprehensive temporal match and enhancing the video retrieval accuracy. 

\section{Model}
\label{sec:v2vir:model}

This paper unveils a different deep learning model tailored for both 2D and 3D similar video retrieval, marking advancements in the field. Our model is founded on a customized autoencoder structure, uniquely incorporating Convolutional LSTM (ConvLSTM), residual connections, and transformer blocks to form an efficient 2D sequence-to-sequence autoencoder.

Within the presented architecture for 2D video retrieval, it is imperative to discern between the established methodologies and novel introductions that are part of our research contribution. Starting off, Block R, which utilizes the ConvLSTM layer, is a traditional approach and is not new. In contrast, we are the first one to propose the LRBP block. The rationale behind adopting a bi-directional version of the ConvLSTM in the LRBP block is to adeptly capture the temporal dynamics in video data, especially those that might possess a reversible nature. Our empirical findings corroborate that the LRBP block manifests a superior performance relative to its non bi-directional counterpart, which is epitomized by the URB block. While the URB and UQB blocks remain a typical technique in employing ConvLSTM for video data, we introduce the amalgamation of the Quasi 4D CNN within an autoencoder structure through the UQB block. This innovation holds promising potential in transforming how visual temporal structures are perceived. Meanwhile, the UTB block, which leverages the transformer for embedding processing, does not offer a fresh perspective within our framework. We are the first to combine the blocks LRBP and UQB in this form. 

\subsection{2D Seq2seq Autoencoder}

\begin{figure}[htbp]
    \centering
    \begin{subfigure}[b]{0.15\textwidth}
            \centering
            \includegraphics[width=1\linewidth]{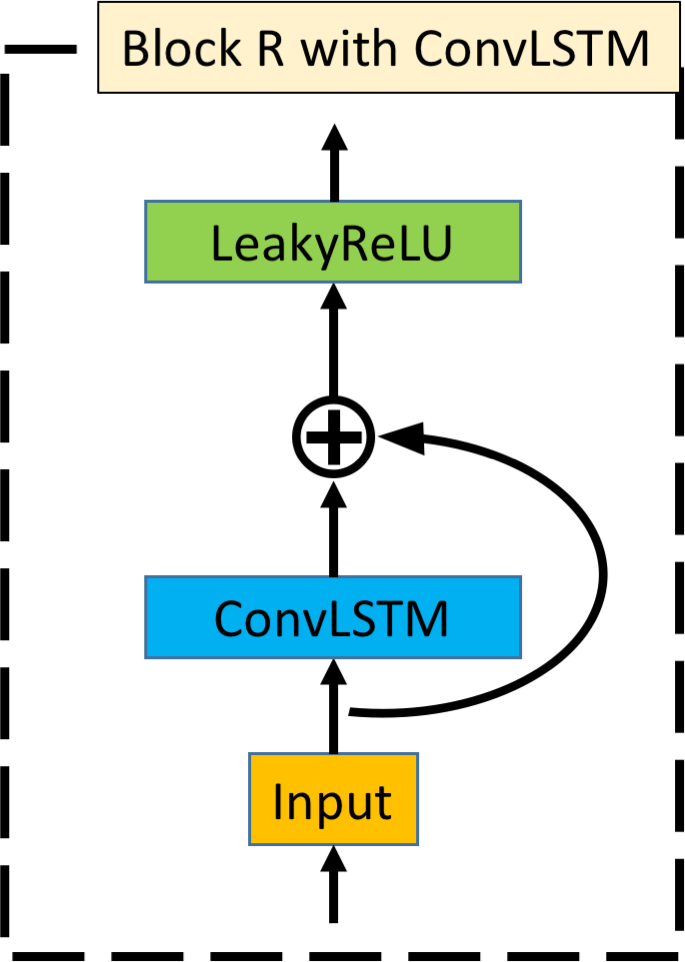}
            \caption{Block R}
    \label{fig:block_r}
    \end{subfigure}
    \begin{subfigure}[b]{0.15\textwidth}
            \centering
            \includegraphics[width=1\linewidth]{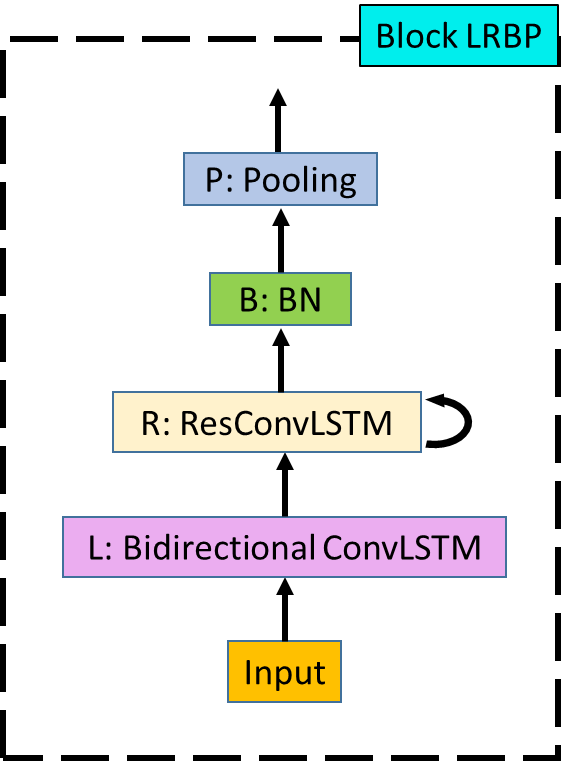}
            \caption{Block LRBP}
    \label{fig:block_lrbp}
    \end{subfigure}
    \begin{subfigure}[b]{0.15\textwidth}
            \centering
            \includegraphics[width=1\linewidth]{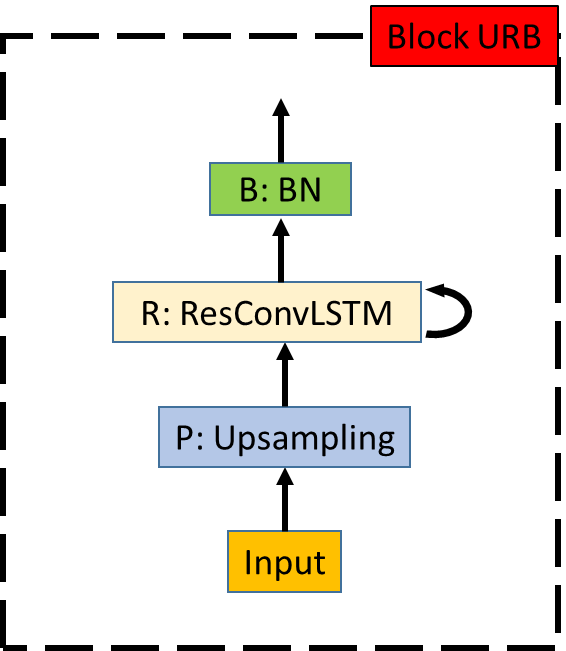}
            \caption{Block URB}
    \label{fig:block_urb}
    \end{subfigure}
    \begin{subfigure}[b]{0.15\textwidth}
            \centering
            \includegraphics[width=1\linewidth]{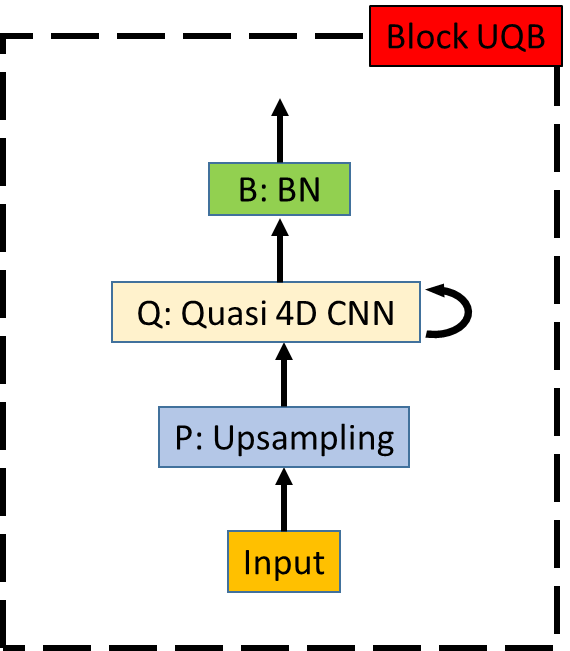}
            \caption{Block UQB}
    \label{fig:block_uqb}
    \end{subfigure}
    \begin{subfigure}[b]{0.15\textwidth}
            \centering
            \includegraphics[width=1\linewidth]{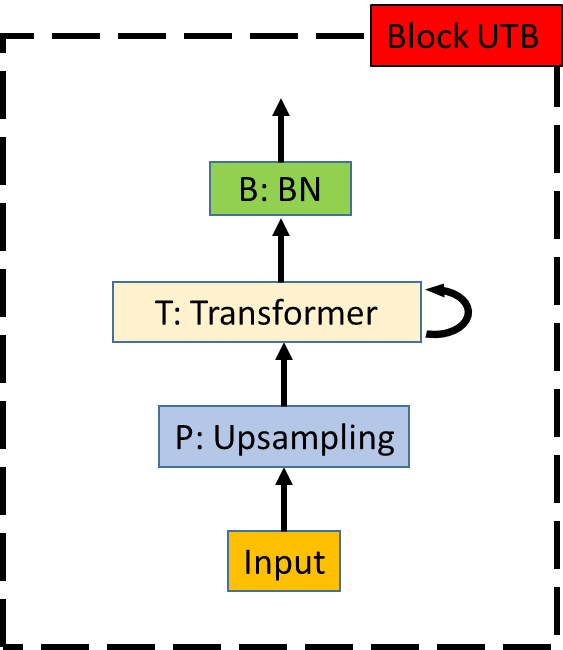}
            \caption{Block UTB}
    \label{fig:block_utb}
    \end{subfigure}
    
    \caption{ 
    These figures show the model components for the 2D cases.
    (\protect\subref{fig:block_r}) Shows the structure of block R. It contains a residual connected ConvLSTM layer and LeakyReLU activation function.
    (\protect\subref{fig:block_lrbp}) Shows the structure of block LRBP. 
    (\protect\subref{fig:block_urb}) Shows the structure of block URB.
    (\protect\subref{fig:block_uqb}) Shows the structure of block UQB.
    (\protect\subref{fig:block_utb}) Shows the structure of block UTB.}
    \label{fig:basic_blocks}
\end{figure}

This paper proposes a 2D sequence-to-sequence autoencoder for similar video retrieval. We first define several basic neural network modules as blocks to simplify the explanations. Block R, shown in Figure \ref{fig:block_r}, consists of a convolutional long short-term memory (ConvLSTM) layer with a residual connection and a LeakyReLU activation function. The LRBP block, defined in \ref{fig:block_lrbp}, connects a block R with a pooling layer and a batch normalization layer. The encoder uses this block to compress the input into a lower dimension. The URB block, defined in \ref{fig:block_urb}, connects an up-sampling layer with block R and a batch normalization layer. The decoder uses this block to restore the embedding to a higher dimension. There is no residual connection between the layers of the encoder and decoder, which ensures the embedding vector after the encoder is a representation of the input video. To improve the generalization of the proposed model and enhance training speed, we include a residual connection within each block of LRBP and URB.

In addition to the URB block, we propose two other types of blocks to support the decoder: the UQB block and the UTB block. These blocks are used to handle the extra dimensionality in the 3D case. The UQB block replaces the ConvLSTM layer in the URB block with a Quasi 4D CNN layer, and the UTB block replaces the ConvLSTM layer with a transformer layer.

\begin{figure}[htbp]
    \centering
    \begin{subfigure}[b]{0.23\textwidth}
            \centering
            \includegraphics[width=1\linewidth]{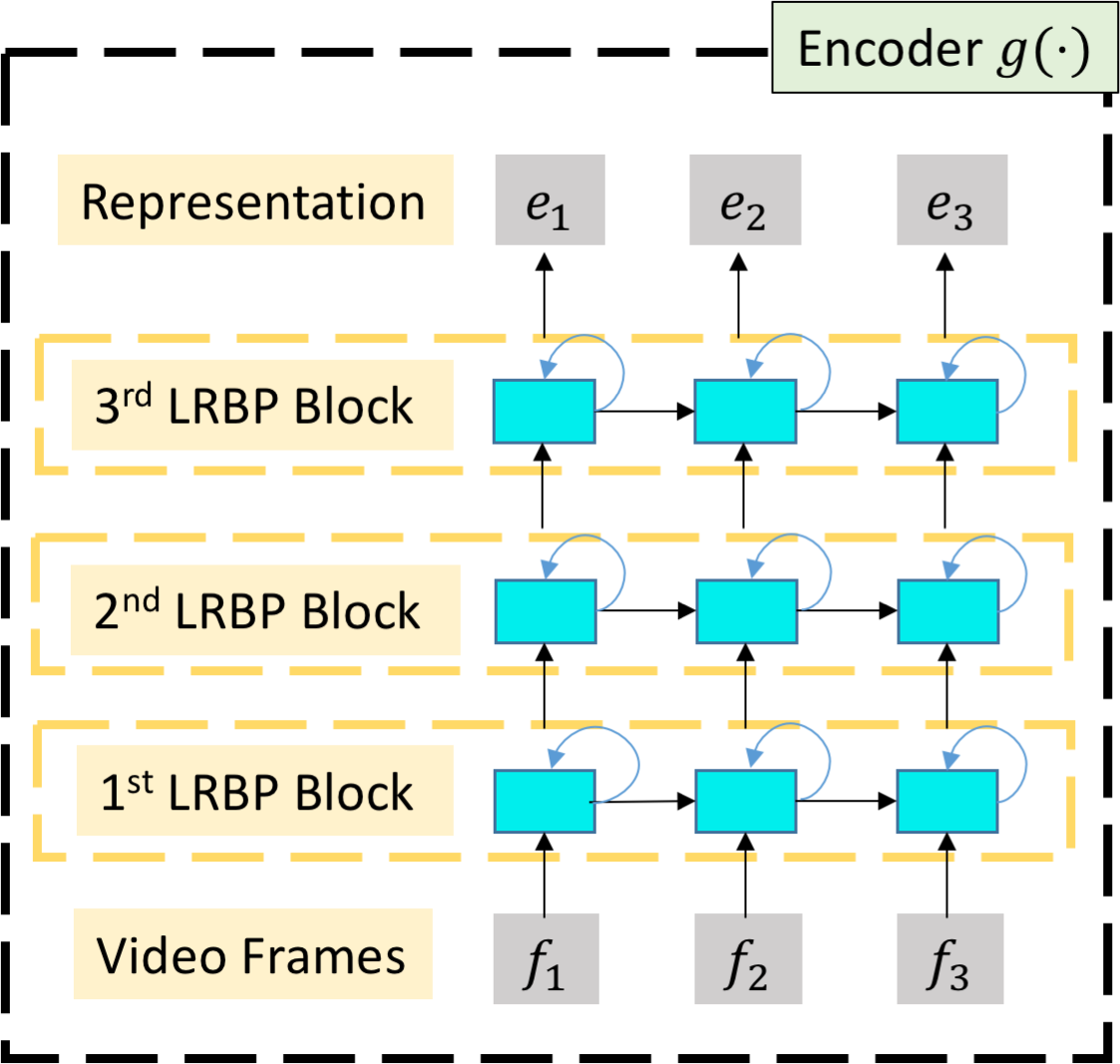}
            \caption{The encoder}
    \label{fig:encoder}
    \end{subfigure}
    \begin{subfigure}[b]{0.23\textwidth}
            \centering
            \includegraphics[width=1\linewidth]{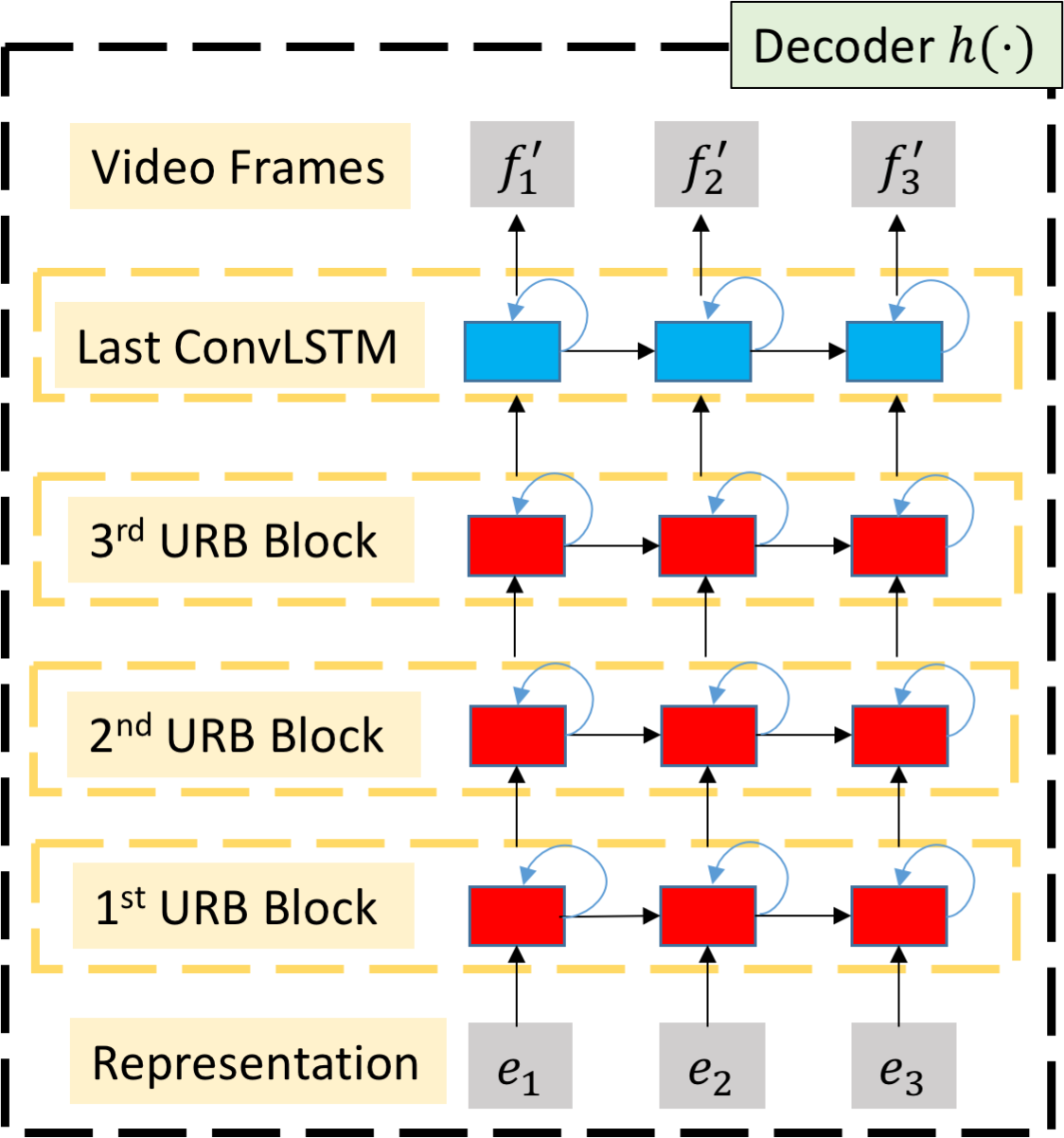}
            \caption{The decoder}
    \label{fig:decoder}
    \end{subfigure}
        
    \caption{(\protect\subref{fig:encoder}) Shows the encoder architecture. (\protect\subref{fig:decoder}) Shows the decoder architecture. The input video has three frames and the representation has three vectors.}
    \label{fig:encoder_decoder}
\end{figure}

\begin{figure}[htbp]
    \centering
    \includegraphics[width=1\linewidth]{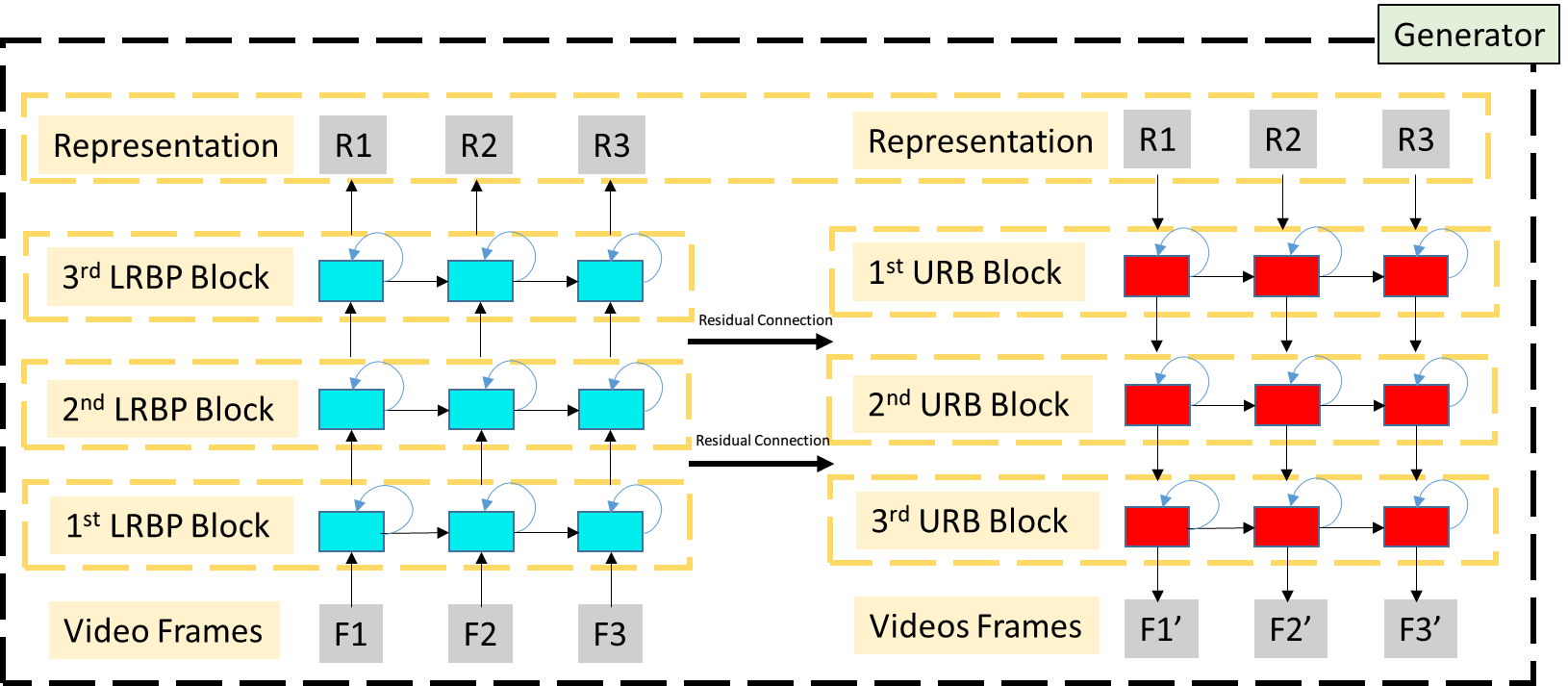}
    \caption{ Illustrates how the encoder and decoder work together as an autoencoder, using LRBP and URB blocks as examples. The number of LRBP and URB blocks can be adjusted for better performance. }
    \label{fig:generator}
\end{figure}

We trained the autoencoder to recover the original video and used the following loss function during training. In the autoencoder architecture, we have used a combination of the LRBP block in the encoder and the URB block in the decoder, as shown in Figure \ref{fig:encoder_decoder}. Moreover, Figure \ref{fig:generator} illustrates the comprehensive function of the proposed autoencoder architecture

\begin{equation}
L_{\text{autoencoder}} =  \mathcal{L}\left(v, h_{\tau}(f_{\theta}(v))\right)
\label{eqn-autoencoder-loss}
\end{equation}
where $ \textbf{v} $ is an input video, $ f_{\theta}(\cdot) $ is the encoder, $ h_{\tau}(\cdot) $ is the decoder and $ \mathcal{L}(\cdot, \cdot) $ represents a loss function. Hence $ h_{\tau}(f_{\theta}(v)) $ is the recovered video generated by the autoencoder.

In our 2D video retrieval architecture, the LRBP and UQB blocks stand out as our primary contributions. The LRBP block employs an advanced bi-directional ConvLSTM, adeptly capturing intricate video dynamics. In contrast, the UQB block, incorporating its Quasi 4D CNN, brings forward a unique representation technique. This UQB innovation is not limited to 2D but will be expanded upon in 3D contexts. Collectively, these innovations significantly enhance video embedding, optimizing the retrieval process.

\subsection{3D Seq2seq Autoencoder}

We also propose a 3D sequence-to-sequence autoencoder, an extension of the 2D sequence-to-sequence autoencoder model, which can be trained unsupervised for pretraining and fine-tuning with a supervised dataset. We are going to define three different variants for the proposed 3D model, namely M1-3D, M2-3D, and M3-3D.

In Table \ref{table-model-architecture}, we describe the 2D and 3D architectural differences among the six proposed 2D and 3D models. This block-based framework allows flexibility in the architecture and enables us to experiment with different combinations of blocks to achieve the best performance for similar video retrieval. The major difference between these models is: M1 is the 2D baseline model, M2 adds a Quasi 4D CNN layer in the decoder, M3 adds a transformer unit in the decoder; M1-3D is the 3D baseline model using a 3D ConvLSTM cell in the autoencoder's encoder and decoder parts. This allows the model to effectively extract the important features from the 3D video data and convert them into a compact and informative embedding for similar video retrieval. The second variant, M2-3D, replaces the 3D ConvLSTM cell with a Quasi 4D CNN layer for handling the 3D video reconstruction. The last variant, M3-3D, uses a transformer unit in the decoder.

\begin{figure}[htbp]
    \centering
    \begin{subfigure}[b]{0.15\textwidth}
            \centering
            \includegraphics[width=1\linewidth]{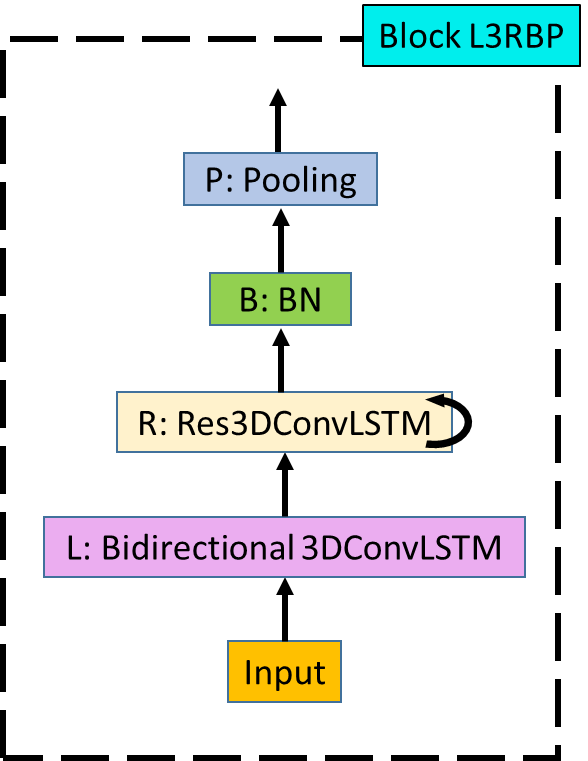}
            \caption{Block L3RBP}
    \label{fig:block_l3rbp}
    \end{subfigure}
    \begin{subfigure}[b]{0.15\textwidth}
            \centering
            \includegraphics[width=1\linewidth]{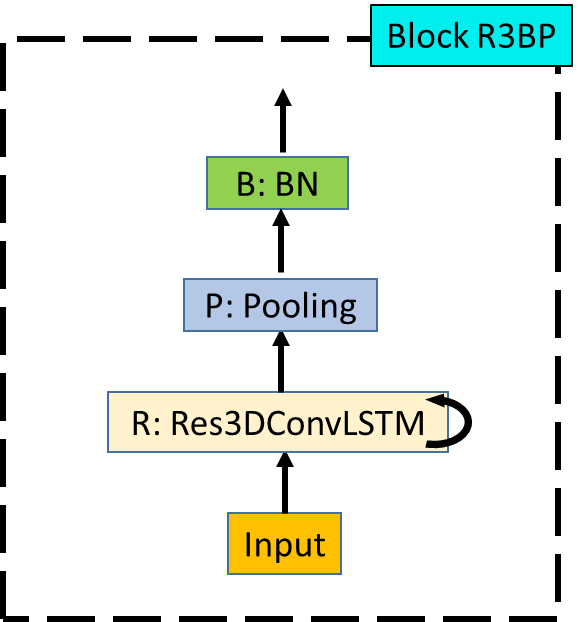}
            \caption{Block R3BP}
    \label{fig:block_r3bp}
    \end{subfigure}
    \begin{subfigure}[b]{0.15\textwidth}
            \centering
            \includegraphics[width=1\linewidth]{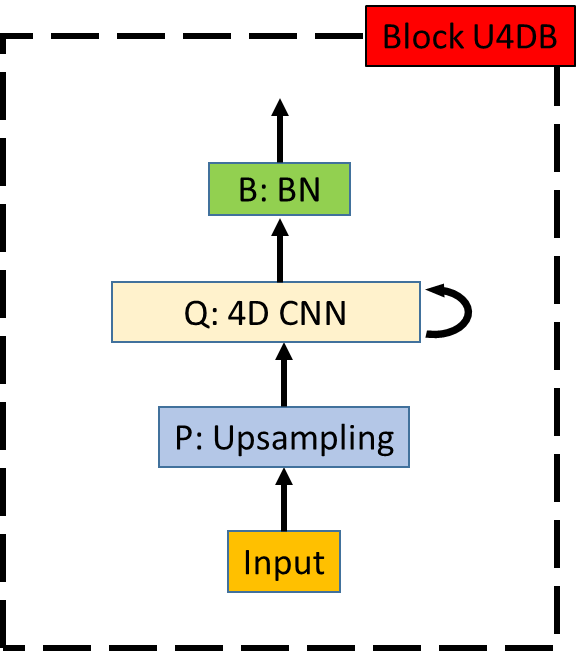}
            \caption{Block U4DB}
    \label{fig:block_u4db}
    \end{subfigure}
    \caption{ 
    These figures show the model components for the 3D cases.
    (\protect\subref{fig:block_l3rbp}) Shows the structure of block L3RBP.
    (\protect\subref{fig:block_r3bp}) Shows the structure of block R3BP. 
    (\protect\subref{fig:block_u4db}) Shows the structure of block U4DB.}
    \label{fig:basic_blocks_3d}
\end{figure}

\begin{table}[ht]
\caption{Comparison of encoder and decoder blocks in all models}
\label{table-model-architecture}
\begin{center}
\begin{scriptsize}
\begin{sc}
\begin{tabular}{lclll}
\toprule
Model No. & 2D/3D & Encoder & Decoder & Diff \\
\midrule
M1    & 2D  & 3 LRBP & 3 URP     & 2D Baseline\\
M2    & 2D  & 3 LRBP & 3 UQP     & \quad + Quasi 4D CNN\\
M3    & 2D  & 3 LRBP & 3 UTP     & \quad + Transformer \\
\midrule 
M1-3D & 3D  & 3 L3RBP & 3 R3BP   & 3D Baseline\\
M2-3D & 3D  & 3 L3RBP & 3 U4DB   & \quad + Quasi 4D CNN\\
M3-3D & 3D  & 3 L3RBP & 3 UTP    & \quad + Transformer\\
\bottomrule
\end{tabular}
\end{sc}
\end{scriptsize}
\end{center}
\vskip -0.1in
\end{table}

For the 3D models, a series of unique building blocks are employed: L3RBP, R3BP, and U4DB. The L3RBP is our signature contribution, introducing a bi-directional connection to the 3D ConvLSTM. This innovative structure is tailored explicitly for video inquiry, and it stands as a natural 3D progression of the LRBP block that we pioneered for 2D scenarios within this paper. On the other hand, the R3BP block harnesses the 3D ConvLSTM layer in a manner that aligns more with conventional techniques of video data processing. The U4DB block, conceptualized as the 3D iteration of the UQB block, primarily hinges on the 4D CNN layer. While the underpinning 4D CNN mechanism has been explored by others in the realm of video data, our approach offers a unique blend that synergizes seamlessly with our overarching model architecture. We are the first to combine the block L3RBP in such a form.

In the realm of 3D video retrieval, the new models introduce innovative adaptations that advance the field in two significant ways. Firstly, compared to 2D models, the 3D models harness the added depth dimension of videos to provide richer and more accurate representations of their content. This is achieved via a novel incorporation of 3D Convolutional LSTM cells, 3D CNN and Quasi 4D CNN layers into the autoencoder's encoder and decoder structure. The adaptation captures temporal correlations across frames more efficiently, leading to a more informative embedding for video similarity retrieval.

The second innovation is a distinct approach in dealing with the challenges posed by other state-of-the-art 3D video retrieval methods. While many existing techniques struggle to maintain performance with increasing video dimensionality, the proposed models, especially M2-3D and M3-3D, demonstrate robustness in handling higher-dimensional data. M2-3D substitutes the 3D ConvLSTM cell with a Quasi 4D CNN layer, introducing a novel way to manage 3D video reconstruction. The M3-3D model introduces a transformer unit in the decoder, a significant leap that helps model long-range temporal dependencies, offering superior performance in similarity retrieval tasks.

Conclusively, these enhancements furnish our 3D models with the capability to set pioneering standards for similar video retrieval. Their robust architecture and performance not only outshine the 2D models but also firmly position them at the forefront, rivaling other contemporary methods in the domain.

\section{Algorithm}

This section presents the problem setting of our video embedding learning problem. We define a distance metric for video embeddings that a triplet loss can train. We then propose the bi-directional dynamic time warping algorithm to convert the embedding into a distance metric.

\subsection{Problem Setting}

We begin with addressing the problem of learning a pairwise similarity
function for similar video retrieval from the relative information of pair/triplet-wise video relations. For a given query video and a set of candidate videos, the goal is to compute the similarity between the query and every candidate video and use it for ranking the entire set of candidates in the hope that similar videos are retrieved at the top ranks. To formulate this process, we define the similarity between two arbitrary video clips $q$ and $p$ as the squared Euclidean distance in the video embedding
space.

\begin{equation}
\label{distance-definition}
D(f_{\theta}(q), f_\theta(p)) = \lVert f_\theta(q) - f_\theta(p) \rVert^2_2
\end{equation}
where $f_\theta(\cdot)$ is the embedding function that maps a video to
a point in a Euclidean space, and $\theta$ are the system parameters.
Additionally, we define a pairwise indicator function $I(\cdot, \cdot)$, specifying whether a pair of videos is near-duplicated. Formally, $I(q, p) = 1$ if $q, p$ are NDVs (near-duplicate videos) and $0$ otherwise.

The proposed model's objective is to learn an embedding function $f_\theta(·)$
that assigns smaller distances to similar video pairs compared to
non-similar ones. Given a video with feature vector $v$, a
similar video with $v+$ and a dissimilar video with $v-$, the embedding function $f_\theta(\cdot)$ should map video representations to a common space $R^d$, where $d$ is the dimension of the feature
embedding, in which the distance between query $v$ and positive
$v+$ is always smaller than the distance between the query
$v$ and negative $v-$:

\begin{equation}
\label{distance-property}
D(f_\theta(v), f_\theta(v^+)) < D(f_\theta(v), f_\theta(v^-)),
\end{equation}
where $I(v, v^+)=1$ and $I(v, v^-)=0$ for all $v, v^+$ and $v^-$.

\subsection{Triplet Loss}

\begin{figure}[ht]
\begin{center}
\centerline{\includegraphics[width=.8\columnwidth]{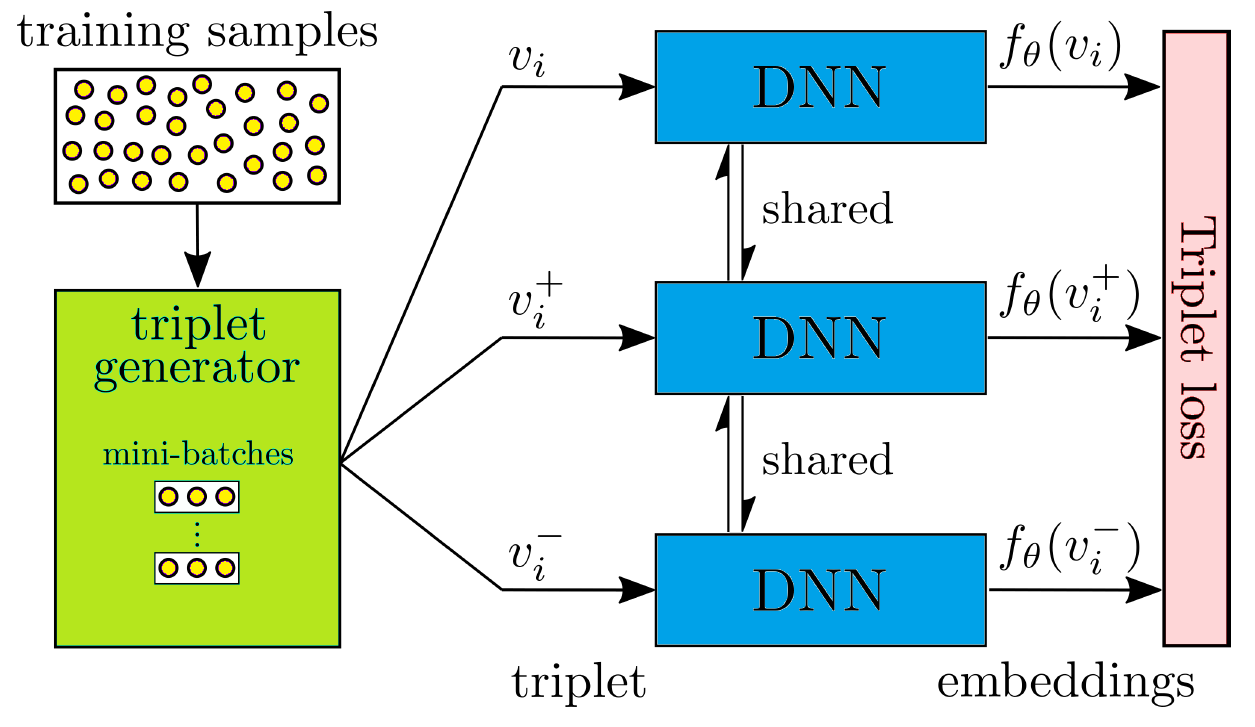}}
\caption{Illustration of a triplet loss in the proposed framework.}
\label{fig-triplet-loss}
\end{center}
\end{figure}

We use triplet loss to train the model and learn the above distance mapping function as a neural network. We define a collection of $N$ training instances in the form of triplets $T = \{(v_i, v_i^+, v_i^-), i=1,...,N\}$ where $v_i, v_i^+, v_i^-$ are feature vectors of a video, a similar positive video clip, and a negative video clip. A triplet expresses a relative similarity order among the three videos. We define the following hinge loss function for a given triplet:

\begin{equation}
\begin{adjustbox}{max width=\linewidth}
$L_\theta(v_i, v_i^+, v_i^-) = \max\{0, D(f_{\theta}(v_i), f_\theta(v_i^+))- D(f_{\theta}(v_i), f_\theta(v_i^-)) + \tau\}$
\end{adjustbox}
\label{triplet-loss-distance}
\end{equation}

where $\tau$ is a margin parameter to ensure a sufficiently large difference between the positive and negative distances. This margin parameter also affects how the model is penalized if there is a violation for the desired triplet distance property. Finally, we use batch gradient descent to optimize the objective function described as triplet loss

\begin{equation}
\label{eqn-triplet-loss}
\min_\theta \sum_{i=1}^{N} L_\theta(v_i, v_i^+, v_i^-) + \lambda \lVert \theta \rVert^2_2
\end{equation}

where $\lambda$ is a regularization parameter to prevent over-fitting of the model, and $N$ is the total size of a triplet mini-batch. This triplet loss is also visualized in Figure \ref{fig-triplet-loss}. Minimizing this loss should narrow the query-positive distance while widening the query-negative distance, and thus lead to a representation satisfying the desirable ranking order. With an appropriate triplet generation strategy in place, the model should eventually learn a video representation that improves the effectiveness of the relevant video retrieval solution.

\subsection{Training Scheme}


We assume we have a training set of videos $\{v_i | i\in X\}$ together with object classification already performed on a subset $X_c \subset X$. To this end, each $\{v_i | i\in X_c\}$ has the corresponding class. As described in Algorithm \ref{alg:model_pretraining_method}, we use a four-step training scheme to optimize the performance of the proposed autoencoder-based video retrieval model. In step 2, a single sample $i \in X_c$ can yield multiple triplets and thus the need for $ \bar{X_c}$. Step 4 intentionally increases the sample number of under-studied data to encourage the model to learn better on the difficult part of the dataset. The output of the algorithm is the semi-supervised trained encoder.

\begin{algorithm}[tb]
   \caption{Training process for proposed model}
   \label{alg:model_pretraining_method}
 \hspace*{\algorithmicindent} \textbf{Input}: Training data $\{v_i | i\in X\}$
\begin{algorithmic}[1]
 \State Pre-train the autoencoder large video dataset from scratch with autoencoder loss (\ref{eqn-autoencoder-loss}) on $\{v_i | i \in X\}$.
 \State Train the autoencoder with triplet-loss (\ref{eqn-triplet-loss}) on the training set $\{(v_i, v_i^+, v_i^-) | i \in \bar{X_c}\}$, where $v_i$ and $v_i^+$ are from the same classes and $v_i^-$ is from a different class. Here $\bar{X_c}$ is a multiset with respect to $X_c$.
 \State Compute the similarity measure (\ref{distance-definition}) for the training set. For each sample $i \in X_c$, find the top 20\% training samples with highest $L_\theta(v_i, v_i^+, v_i^-)$ values. Let us denote these challenging samples as $\{(v_i, v_i^+, v_i^-) | i \in \tilde{X_c}\}$.
 \State Fine tune the autoencoder with triplet-loss (\ref{eqn-triplet-loss}) and train with 50\% of samples from the challenging sample sets $\tilde{X}_c$ and 50\% of samples from the general training sets $\bar{X_c} \slash \tilde{X}_c$. 
\end{algorithmic}
\end{algorithm}


\subsection{Video Similarity Search}

Note that each video is represented as a sequence of frames or clips. We denote a video as $v$, and each frame as $z_i$, where $i \in M$ and $M$ is the number of frames in the video. Alternatively, we can represent a video as a sequence of clips, denoted as $\textbf{p}_i$, where $i \in N_x$ and $N_x = \lfloor \frac{M}{k} \rfloor$. The clips can be formed in two ways: disjoint consecutive frames or overlapping consecutive frames.

We then compute an embedded representation vector, $v^p$, for each clip of video $v$ by mapping it to a vector using the encoder. We use a dynamic time-warping (DTW) method to measure similarity between two videos. This method is commonly used in time series analysis to measure the similarity between two temporal sequences. The DTW distance between video clips $v_1$ and $v_2$ is solved as a dynamic programming problem, with the core state transfer formula being

\begin{equation}
D_{i,j} := \lVert v_1^i - v_2^j \rVert^2_2 + min( D_{i-1,j}, D_{i,j-1}, D_{i-1,j-1}).
\end{equation}

In order to provide a comprehensive solution to the video embedding matching problem, we expand on the traditional dynamic time warping (DTW) algorithm by proposing the bi-directional dynamic time warping (Bi-DTW) algorithm. Bi-DTW is distinguished by its ability to execute matching from both forward and backward directions, leading to an enhancement in the accuracy of the video retrieval system. The Bi-DTW algorithm computes both $DTW(v_1,v_2)$ and $DTW(\text{reverse}(v1), \text{reverse}(v2))$ where $\text{reverse}(v)$ is the sequence of clips in the reverse order. The output of Bi-DTW algorithm is  $min(DTW(v1,v2), DTW(\text{reverse}(v1), \text{reverse}(v2)))$. Hence, Bi-DTW, with its dual-directional operation, not only enables us to determine the degree of similarity between two videos more effectively, but also enhances the ranking process of candidate videos. By comparing relevance from both the original and reverse sequences of the query video, it ensures a more comprehensive matching, thereby providing a superior and more versatile solution for the video retrieval task.

\section{Numerical Experiments}

In this section, we present experimental results, which demonstrate the effectiveness of the proposed method in terms of retrieval accuracy and computational efficiency when compared to the state-of-the-art methods.

\subsection{Datasets}

We have conducted numerical experiments using the CCWebVideo, YouTube-8M-sub, s3DIS, and Synthia-SF public datasets to evaluate the performance of the proposed model. The CCWebVideo dataset includes 12,790 videos and 27\% near-duplicates. The YouTube-8M dataset is a large-scale video dataset which includes more than 7 million videos with 4,716 classes. The YouTube-8m-sub dataset we use is a random subset of YouTube-8m with around 100 videos from each class of YouTube-8m datasets. The Stanford 3D Indoor Scene (S3DIS) dataset contains 6 large-scale indoor areas with 271 rooms. Each point in the scene point cloud is annotated with one of the 13 semantic classes. The Synthia dataset is a synthetic dataset that consists of 9,400 multi-viewpoint photo-realistic frames rendered from a virtual city and comes with pixel-level semantic annotations for 13 classes. Synthia-SF is a subset of Synthia that only covers San Francisco. Table \ref{table-datasets} summarizes the number of samples in each dataset. For the 3D datasets, we concatenate the depth information as an additional input dimension to the proposed 3D video auto-encoder. 

\begin{table}[H]
\caption{Descriptions for all four datasets.}
\label{table-datasets}
\vskip 0.15in
\begin{center}
\begin{scriptsize}
\begin{sc}
\begin{tabular}{lrrrr}
\toprule
         &               &  No. of    &  No. of & Avg.  \\
 Dataset   &  Type       & classes    &  videos & video \\
        &                &            &         & length \\        
\midrule
CCWebVideo     & 2D & 24   & 533 & 265s \\
YouTube-8m-sub & 2D & 1000 & 100 & 3000s \\
S3DIS          & 3D & 6  & 1600 & 251s \\
Synthia-SF     & 3D & 6  & 417 & 83s \\
\bottomrule
\end{tabular}
\end{sc}
\end{scriptsize}
\end{center}
\vskip -0.1in
\end{table}

\subsection{Implementation}

We implemented the model using the Tensorflow 1.15 framework, and trained it on NVIDIA 3070 GPUs or equivalents. The model was trained with minibatch sizes of 32 and 8 clips for the 2D and 3D datasets respectively, using 4 GPUs in parallel. We employed the softmax loss function for the auto-encoder and applied L2 regularization of penalty ratio of 0.001 to the model's trainable parameters. These parameters were initialized using Xavier initialization.

The model was optimized using stochastic gradient descent (SGD) with an initial learning rate of 0.001 which decayed by 10 times every 10 epochs. Training ran for 50 epochs with early stopping after 5 epochs of no improvement. L2 regularization of 0.001 and momentum of 0.9 were used.

For the seq2seq auto-encoder, sequence-wise normalization was applied across multiple video sequences within each mini-batch. We computed the mean and variance statistics across all timesteps within this mini-batch for each output channel. Activation functions are ReLU. 

For the encoder, it contains 3 LRBP layers and a dense layer. Each LRBP block took an input tensor of size 256x256x3. The BidirectionalConvLSTM layer employed a 3x3 kernel, a stride of 1, and had 64 hidden states. The subsequent ResConvLSTM used the same kernel size, padding, and stride but contained 32 hidden states. The output tensor after the residual connection was of size 256x256x96, which was then reduced to 128x128x16 after pooling. The output tensor was further reduced to 32x32x16 after the third LRBP block. This tensor was mapped to a 4000-entry embedding vector by a dense layer. In the Quasi 4D CNN, we used a 3x3x3 kernel. The transformer was applied to the embedding vectors, and consisted of 5 self-attention layers each with 3 attention heads. The transformer's hidden size was 512, and its intermediate size was 2048.

For the URB decoder, it contains three URB layers. The embedding vector was first transformed by a dense layer to 32x32x16 before entering the first URB block. Each URB block's ResConvLSTM utilized a 3x3 kernel, a stride of 1, and had 32 hidden states. The final ConvLSTM layer in the decoder had 3 filters, producing an output tensor of size 256x256x3.

The raw videos varied in length and resolution. We normalized them to 30 FPS, resized to 256 height keeping aspect ratio, cropped the center 256x256 region to get square RGB frames, and standardized the values to have 0 mean and unit variance across all videos. This standardized the data in terms of frame size, rate, and value ranges to effectively train deep learning models on the dataset.

\subsection{Benchmarks}

For the 2D video retrieval experiments, we compare the proposed models with a non-deep learning-based baseline method, which uses Scale Invariant Feature Transform (SIFT) to extract features from each frame in a clip and measures similarity between query clips and candidate clips using a bag of words method. Additionally, we also compare the proposed models to the state-of-the-art deep learning-based benchmark algorithm, NDVR-DML \cite{kordopatis-zilos_near-duplicate_2017}, which utilizes a Convolutional Neural Network and a Deep Metric Learning (DML) framework to generate discriminative global video representations and approximates an embedding function for accurate distance calculation between near-duplicate videos.

For the 3D video retrieval experiments, we use an autoencoder as the benchmark model, with 3D CNN layers as the encoder and decoder. This 3D autoencoder is a deep learning-based model that effectively extracts important features from the 3D video data and converts it into a compact and informative embedding for similar video retrieval. The architecture of this 3D autoencoder is similar to the proposed 3D models, with the main difference being the use of 3D CNN layers as opposed to the 3D ConvLSTM cell, Quasi 4D CNN layer, or transformer unit used in the proposed models.

\subsection{Evaluation Metrics}

For evaluation, we use Mean Average Precision (mAP) 
to measure the quality of the retrieval results. Measure mAP is a commonly used evaluation metric in information retrieval. It calculates the average precision of the retrieved results for a given query and is based on a binary relevance scale of 0 (non-relevant) or 1 (relevant).

The mean average precision is formally stated as $mAP = \dfrac{1}{n} \sum_{i=0}^{n}\dfrac{i}{r_i}$, where $n$ is the number of relevant videos to the query video, and $r_i$ is the rank of the $i$th retrieved relevant video \cite{Wu2007}.

In order to evaluate the effectiveness of the proposed video retrieval methods, we carefully constructed a test set by randomly cropping the original training videos and concatenating the new clips. The test queries were formed by randomly selecting a set of frames from a video in the training set, with the original video serving as the query's ground truth. To ensure that the test queries differed from the database records, we designed the test set generation process to include consecutive frames and gaps between frames in the queries. This method of test set generation allows us to simulate real-world scenarios where the input query may be similar but not identical to the database records. The evaluation criteria include successful retrieval by clip and class, with the latter being a more stringent measure of performance.

\subsection{Process}
Training contains four steps, as described in Algorithm \ref{alg:model_pretraining_method}, which optimizes an autoencoder-based video retrieval model's performance. This schema includes pre-training the autoencoder on a large video dataset, training with a triplet-loss on a multiset derived from the training set, computing the similarity measure, and finally, fine-tuning with a second round of triplet-loss training on a mix of challenging and general training sets.

For inference, the dataset is split 70/15/15 into train, validation and test sets. Model efficacy is evaluated under two scenarios: by class and by clip. For class-level evaluation, validation clips are encoded into embeddings and compared against train embeddings. If retrieved training videos match the class of the query, precision is 1. For clip-level, precision is 1 only if the matched video contains the exact query clip. This matching process is repeated for all validation set videos and compared to the training set to compute metrics like mean average precision.

\section{Results}

In this section, we delve into the experimental results of our proposed models applied to 2D and 3D datasets, comparing their performances against various benchmarks. We use the metric of mean Average Precision to assess the models' effectiveness in video clip retrieval both by clip and class. Furthermore, a series of ablation studies help us understand the impact of different training techniques, loss functions, and the dynamic time warping method on the performance of our proposed models.

\subsection{Results for 2D Datasets}

We report the numerical results as mAP for the 2D datasets in Table \ref{table-2d-results}. It shows that the proposed model has superior performance on the 2D datasets regarding the video clip retrieval accuracy by clip and by class. 

\begin{table}[H]
\caption{mAP results for 2D datasets}
\label{table-2d-results}
\vskip 0.15in
\begin{center}
\begin{scriptsize}
\begin{sc}
\begin{tabular}{lcccc}
\toprule
Model  & \multicolumn{2}{c}{CCWebVideo} & \multicolumn{2}{c}{Youtube-8m-sub}\\
       & by class      & by clip   & by class      & by clip \\
\midrule
SIFT     & 58.8\% & 64.7\% & 27.9\% & 38.9\% \\  
NDVR-DML & 84.4\% & 91.7\% & \textbf{64.8\%} & 75.2\% \\  
M1       & 79.6\%  & 85.3\% & 62.3\%  & 71.8\% \\
M2       & 81.7\%  & 90.4\% & 65.4\%  & 74.9\% \\
M3       & \textbf{85.2\%}  & \textbf{92.1\%} & \textbf{64.8\%}  & \textbf{76.8\%} \\
\bottomrule
\end{tabular}
\end{sc}
\end{scriptsize}
\end{center}
\vskip -0.1in
\end{table}

From Table \ref{table-2d-results}, it can be seen that the proposed models M1, M2, and M3 all perform better than the non-deep learning-based SIFT baseline model, with M3 achieving the highest mean average precision across both datasets and both aggregation methods. Additionally, the models outperform the deep learning-based NDVR-DML benchmark model, with M3 achieving comparable or better results on all metrics. Overall, these results demonstrate the effectiveness of the proposed video retrieval models and their ability to achieve high levels of performance on various datasets and evaluation metrics.

\begin{figure}[htbp]
    \centering
    \begin{subfigure}[b]{0.07\textwidth}
            \centering
            \includegraphics[width=1\linewidth]{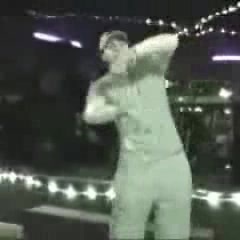}
    \end{subfigure}
    \begin{subfigure}[b]{0.07\textwidth}
            \centering
            \includegraphics[width=1\linewidth]{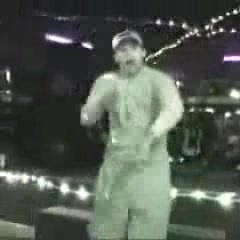}
    \end{subfigure}
    \begin{subfigure}[b]{0.07\textwidth}
            \centering
            \includegraphics[width=1\linewidth]{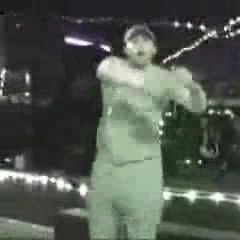}
    \end{subfigure}
    \begin{subfigure}[b]{0.07\textwidth}
            \centering
            \includegraphics[width=1\linewidth]{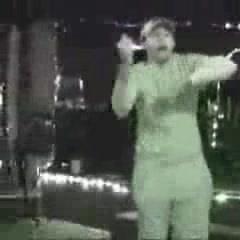}
    \end{subfigure}
    \begin{subfigure}[b]{0.07\textwidth}
            \centering
            \includegraphics[width=1\linewidth]{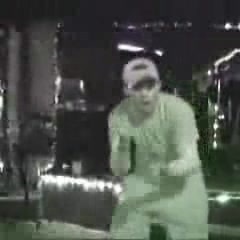}
    \end{subfigure}
    \begin{subfigure}[b]{0.07\textwidth}
            \centering
            \includegraphics[width=1\linewidth]{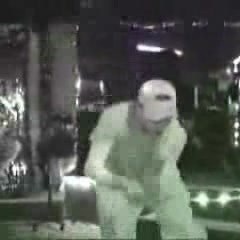}
    \end{subfigure}
    \vspace{-0.1cm}
    \caption{Clip of Comedy class from CCWebVideo that is matched to Music class.}
    \label{fig:example-2d-comedy}
\end{figure}

\begin{figure}[htbp]
    \centering
    \begin{subfigure}[b]{0.07\textwidth}
            \centering
            \includegraphics[width=1\linewidth]{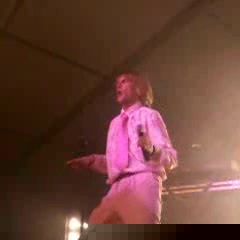}
    \end{subfigure}
    \begin{subfigure}[b]{0.07\textwidth}
            \centering
            \includegraphics[width=1\linewidth]{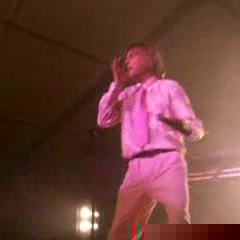}
    \end{subfigure}
    \begin{subfigure}[b]{0.07\textwidth}
            \centering
            \includegraphics[width=1\linewidth]{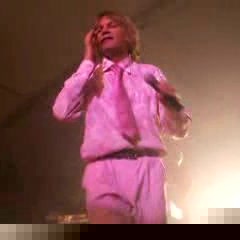}
    \end{subfigure}
    \begin{subfigure}[b]{0.07\textwidth}
            \centering
            \includegraphics[width=1\linewidth]{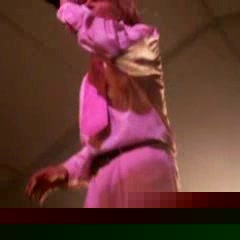}
    \end{subfigure}
    \begin{subfigure}[b]{0.07\textwidth}
            \centering
            \includegraphics[width=1\linewidth]{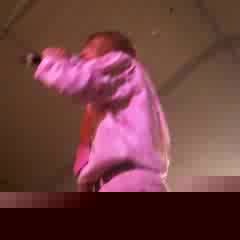}
    \end{subfigure}
    \begin{subfigure}[b]{0.07\textwidth}
            \centering
            \includegraphics[width=1\linewidth]{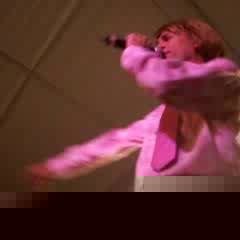}
    \end{subfigure}
    \vspace{-0.1cm}
    \caption{Typical clip from CCWebVideo Music class.}
    \label{fig:example-2d-music}
\end{figure}

\begin{figure}[htbp]
    \centering
    \begin{subfigure}[b]{0.07\textwidth}
            \centering
            \includegraphics[width=1\linewidth]{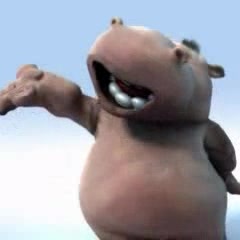}
    \end{subfigure}
    \begin{subfigure}[b]{0.07\textwidth}
            \centering
            \includegraphics[width=1\linewidth]{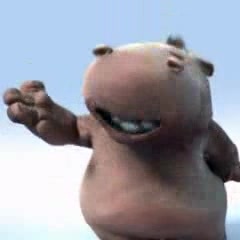}
    \end{subfigure}
    \begin{subfigure}[b]{0.07\textwidth}
            \centering
            \includegraphics[width=1\linewidth]{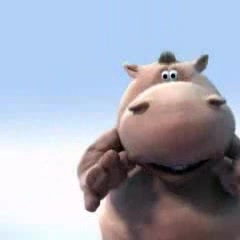}
    \end{subfigure}
    \begin{subfigure}[b]{0.07\textwidth}
            \centering
            \includegraphics[width=1\linewidth]{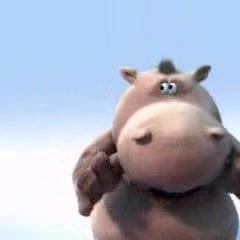}
    \end{subfigure}
    \begin{subfigure}[b]{0.07\textwidth}
            \centering
            \includegraphics[width=1\linewidth]{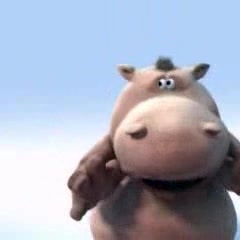}
    \end{subfigure}
    \begin{subfigure}[b]{0.07\textwidth}
            \centering
            \includegraphics[width=1\linewidth]{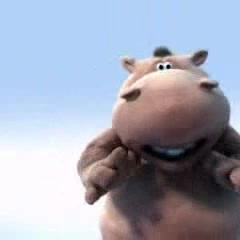}
    \end{subfigure}
    \vspace{-0.1cm}
    \caption{Example clip of Animation class from CCWebVideo.}
    \label{fig:example-2d-animation}
\end{figure}

For the CCWebVideo dataset using model M3, 14.8\% of validation samples failed class-level retrieval. Upon inspection, around 32\% of these failures retrieved videos of the wrong class but with visual similarity. For example, Fig \ref{fig:example-2d-comedy} shows a comedy clip incorrectly matched to the music class. As evident in Figure \ref{fig:example-2d-music}, comedy and music clips can appear visually similar despite different semantic classes. This implies the encoder failed to sufficiently differentiate some inter-class nuances. Introducing the bidirectional dynamic time warping (Bi-DTW) method significantly boosted retrieval accuracy for certain classes. Animation clips improved from 92\% to 98\% class-level mAP, owing to Bi-DTW better handling symmetric motions of characters, as exemplified in Fig \ref{fig:example-2d-animation}.

Overall, the analysis indicates room for improvement in handling inter-class visual similarities. Bi-DTW conferred notable gains for select classes by leveraging bidirectional temporal matching. Further inspection of embedding projections and retrieval results could provide additional insight into model limitations. Targeted sampling and training techniques may help differentiate challenging inter-class pairs.

\vspace{-0.3cm}

\vspace{-0.3cm}

\vspace{-0.3cm}

\subsection{Results for 3D Datasets}

We report the numerical results as mAP for the 3D datasets in Table \ref{table-3d-results}. It shows that the proposed model has superior performance on both 3D datasets about the video clip retrieval accuracy by clip and class. 

\begin{table}[htbp]
\caption{mAP results for 3D datasets}
\label{table-3d-results}
\vskip 0.15in
\begin{center}
\begin{scriptsize}
\begin{sc}
\begin{tabular}{lcccc}
\toprule
Model  & \multicolumn{2}{c}{S3DIS} & \multicolumn{2}{c}{Synthia-SF}\\
       & by class      & by clip   & by class      & by clip \\
\midrule
3D Autoencoder & 58.9\%  & 77.9\%  & 54.5\% & 82.5\% \\
M1-3D          & 61.4\%  & 75.4\%  & \textbf{55.6\%} & 82.1\% \\
M2-3D          & \textbf{62.3\%}  & 73.3\%  & 54.8\% & 81.9\% \\
M3-3D          & 61.1\%  & \textbf{78.1\%}  & 54.5\% & \textbf{83.2\%} \\
\bottomrule
\end{tabular}
\end{sc}
\end{scriptsize}
\end{center}
\vskip -0.2in
\end{table}

According to Table \ref{table-3d-results}, the performance of the proposed models varies across the two datasets and the different levels of aggregation. For example, on the S3DIS dataset, M1-3D and M3 perform similarly well in class-level mAP, with 61.4\% and 61.1\%, respectively, while M3 performs slightly better in terms of clip-level mAP with 78.1\%. On the other hand, on the Synthia-SF dataset, M3 performs slightly better than the other proposed models in both class-level mAP and clip-level mAP.

\begin{figure}[htbp]
\vskip 0.2in
\begin{center}
\centerline{\includegraphics[width=\columnwidth]{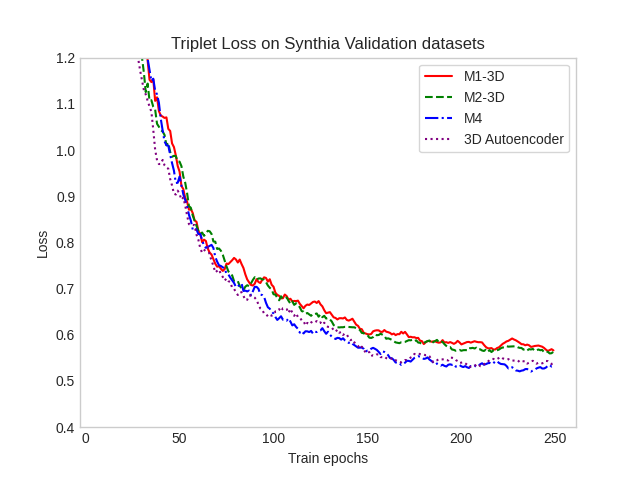}}
\caption{Training loss curve for proposed models}
\label{fig-training-curve-Synthia}
\end{center}
\vskip -0.2in
\end{figure}

The proposed models have not consistently outperformed the 3D autoencoder, except the model M3 is always better or on par with the baseline model. In addition, the table shows that the performance of the models is different when evaluated by class level or clip level. Figure \ref{fig-training-curve-Synthia} shows the training loss when we train the model with the triplet loss on the Synthia dataset. Based on these results, we conclude that the models have great potential to improve the video retrieval results on 3D datasets and that it is important to consider the level of aggregation when evaluating the performance. 

\subsection{Ablation study: Effects of pretraining and challenging sample retraining}

We investigate the impact of two key training techniques on the performance of the proposed deep-learning architecture. Specifically, we study the effects of pretraining on a large and diverse dataset and challenging sample retraining on the accuracy of similar video retrieval. Firstly, we found that pretraining on a diverse dataset such as YouTube-8m-sub is crucial for improving the model's generalization ability and effectively learning from the given training set despite the complexity and intricacies of the video data. This required a significant investment of computation resources, with an approximate 24-hour pretraining period using 4 GPUs.

\begin{table}[htbp]
\caption{Effectiveness of pertaining and challenging sample retraining on CC\_WEB\_VIDEO dataset.}
\label{table-ablation-1}
\vskip 0.15in
\begin{center}
\begin{scriptsize}
\begin{sc}
\begin{tabular}{lccc}
\toprule
Model  & \multicolumn{2}{c}{CC\_WEB\_VIDEO}    \\
       & by class  & by clip  \\
\midrule
M3                  & 54.1\%  & 65.7\% \\
M3 + 2nd pass       & 65.5\%  & 71.3\% \\ 
M3 + challenging 2nd pass  & 69.3\%  & 78.5\% \\
M3 + pre-train      & 84.5\%  & 89.2\% \\
M3 + pre-train      &  & \\
\hspace{2.25em} 
and challenging 2nd pass     & \textbf{85.2\%}  & \textbf{92.1\%} \\
\bottomrule
\end{tabular}
\end{sc}
\end{scriptsize}
\end{center}
\vskip -0.1in
\end{table}

To validate the effectiveness of these techniques, we present an ablation study in Table \ref{table-ablation-1} where we evaluate the performance of different variations of the baseline model, M3, on the CC\_WEB\_VIDEO dataset. The results show that the overall performance of M3 improves significantly when incorporating pretraining and training with challenging samples. Specifically, the model trained with pretraining and challenging samples in the second round achieved the best performance with 85.2\% mAP by class and 92.1\% mAP by clip. This suggests that pretraining and focusing on difficult samples during training can greatly enhance the model's performance for similar video retrieval.

\begin{figure}[htbp]
    \centering
    \begin{subfigure}[b]{0.07\textwidth}
            \centering
            \includegraphics[width=1\linewidth]{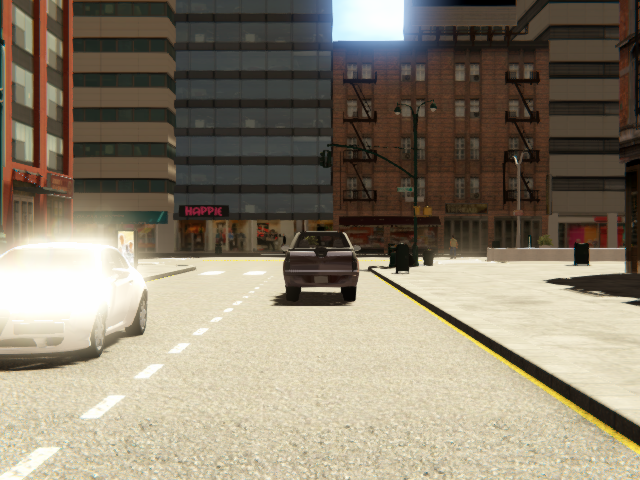}
    \end{subfigure}
    \begin{subfigure}[b]{0.07\textwidth}
            \centering
            \includegraphics[width=1\linewidth]{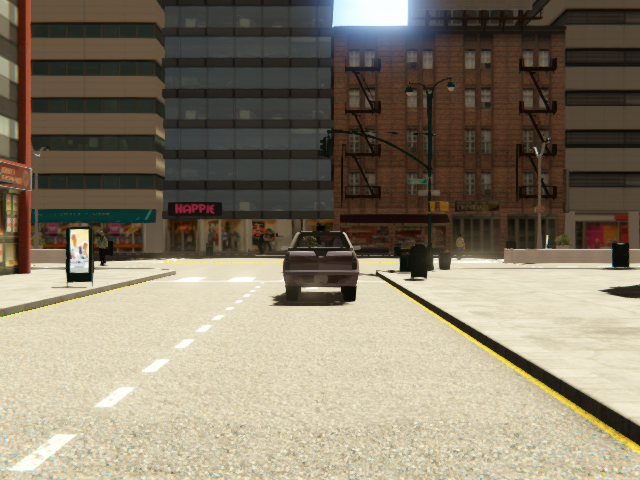}
    \end{subfigure}
    \begin{subfigure}[b]{0.07\textwidth}
            \centering
            \includegraphics[width=1\linewidth]{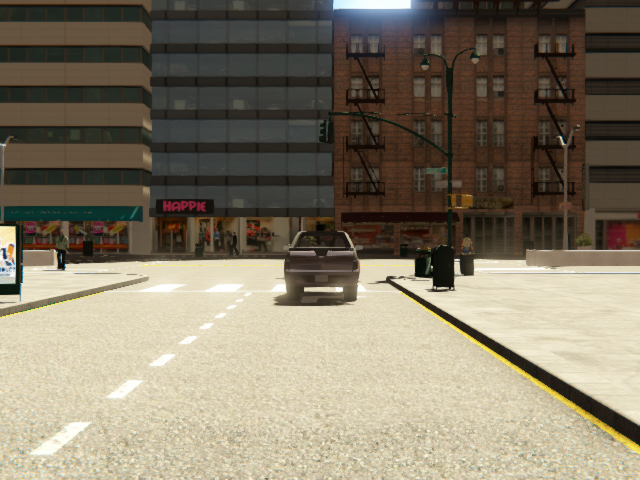}
    \end{subfigure}
    \begin{subfigure}[b]{0.07\textwidth}
            \centering
            \includegraphics[width=1\linewidth]{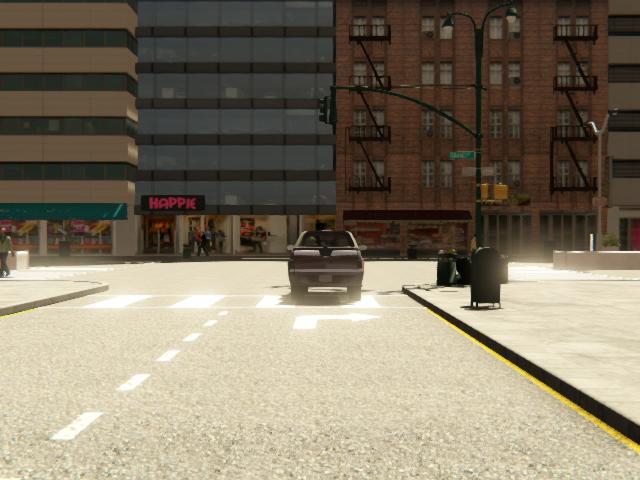}
    \end{subfigure}
    \begin{subfigure}[b]{0.07\textwidth}
            \centering
            \includegraphics[width=1\linewidth]{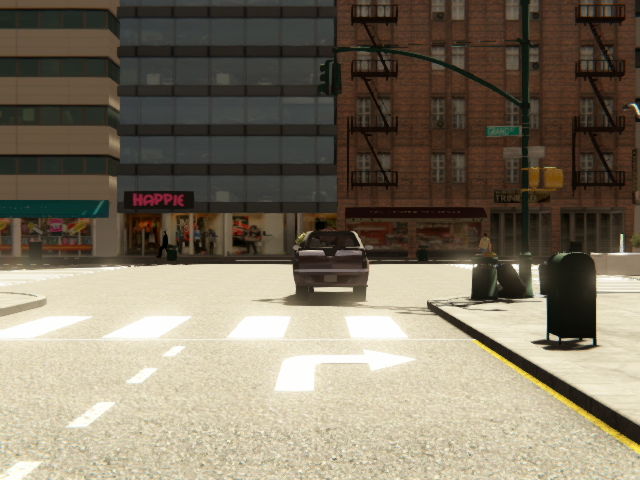}
    \end{subfigure}
    \begin{subfigure}[b]{0.07\textwidth}
            \centering
            \includegraphics[width=1\linewidth]{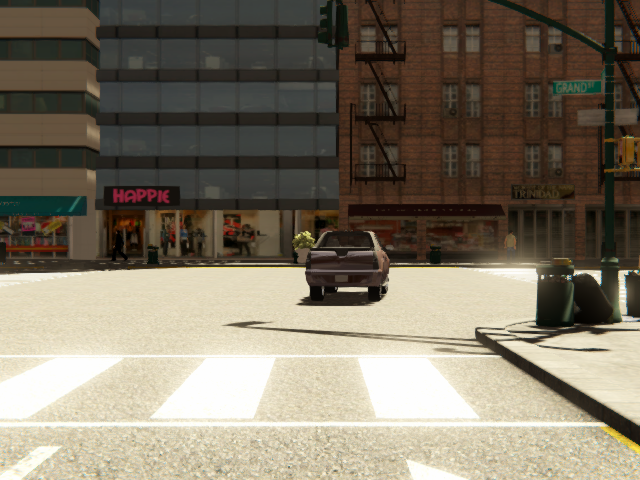}
    \end{subfigure}
    \vspace{-0.1cm}
    \caption{An example clip from Synthia.}
    \label{fig:example-3d-example1}
\end{figure}

\begin{figure}[htbp]
    \centering
    \begin{subfigure}[b]{0.07\textwidth}
            \centering
            \includegraphics[width=1\linewidth]{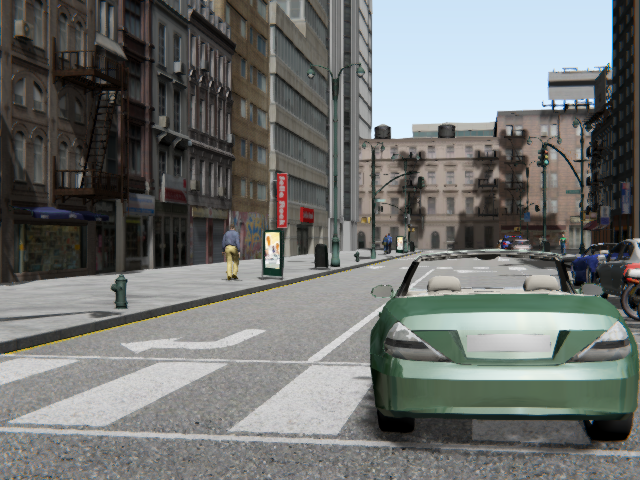}
    \end{subfigure}
    \begin{subfigure}[b]{0.07\textwidth}
            \centering
            \includegraphics[width=1\linewidth]{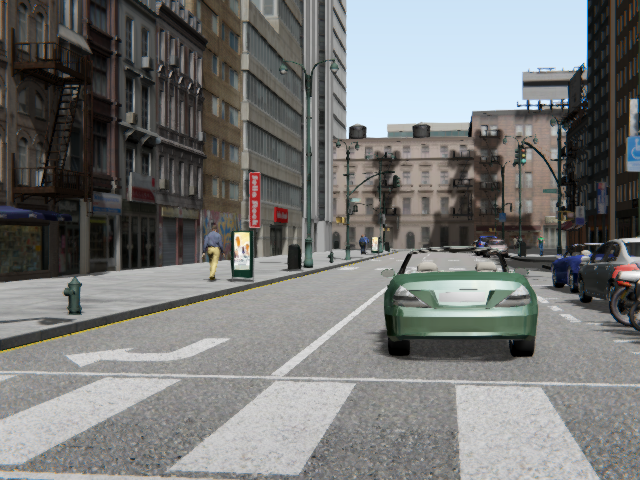}
    \end{subfigure}
    \begin{subfigure}[b]{0.07\textwidth}
            \centering
            \includegraphics[width=1\linewidth]{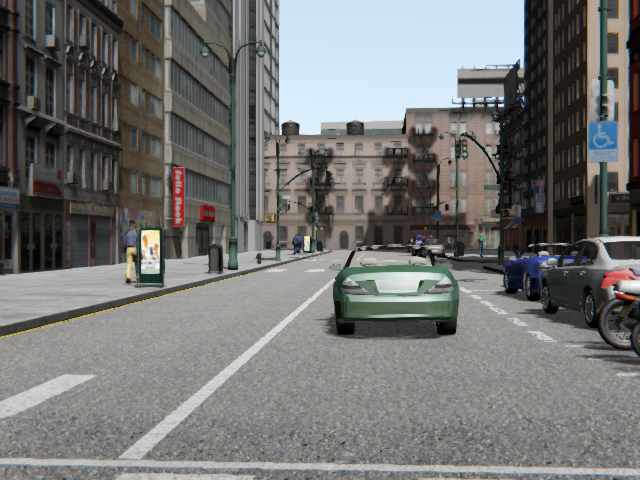}
    \end{subfigure}
    \begin{subfigure}[b]{0.07\textwidth}
            \centering
            \includegraphics[width=1\linewidth]{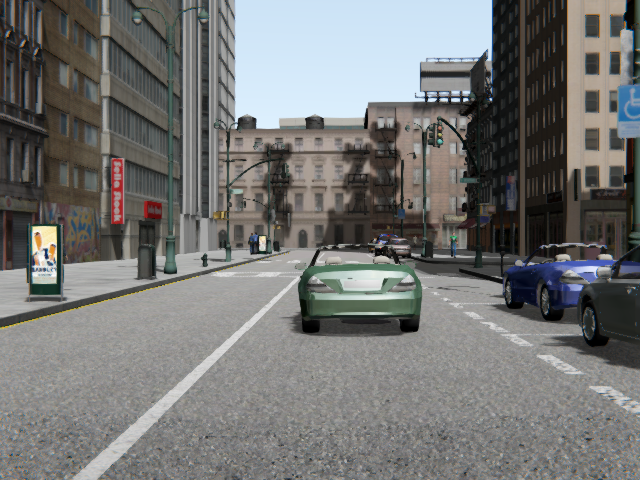}
    \end{subfigure}
    \begin{subfigure}[b]{0.07\textwidth}
            \centering
            \includegraphics[width=1\linewidth]{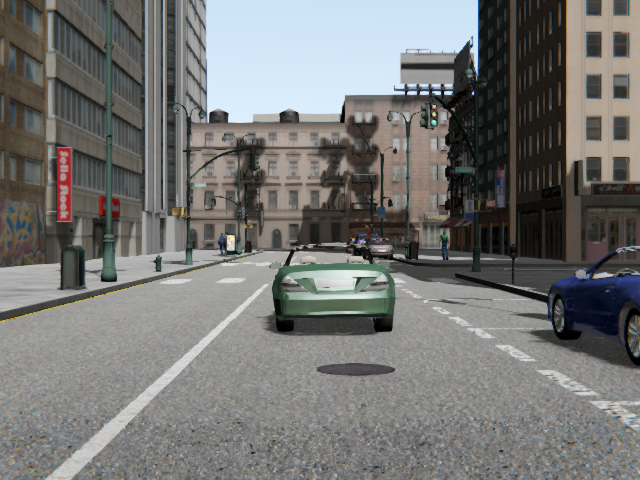}
    \end{subfigure}
    \begin{subfigure}[b]{0.07\textwidth}
            \centering
            \includegraphics[width=1\linewidth]{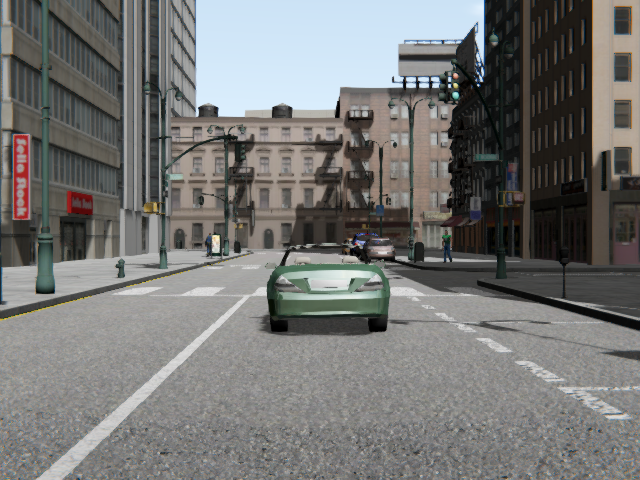}
    \end{subfigure}
    \vspace{-0.1cm}
    \caption{A clip of same location and different weather from Synthia.}
    \label{fig:example-3d-example2}
\end{figure}

Secondly, For the Synthia 3D dataset, Figure \ref{fig:example-3d-example1} shows an example query not improved by Bi-DTW, incorrectly matched to Figure \ref{fig:example-3d-example2} of a different class. Although many Synthia classes depend on filming time or weather, this failure stems from visual similarity of the street scenes. The Bi-DTW method may incorrectly relate clips appearing to drive in reverse on the same road despite different classes. Visually analogous roads filmed in opposing directions can be erroneously aligned by Bi-DTW. Further inspection into geometrically similar backgrounds causing inter-class confusion is needed. Additional constraints or tweaks to Bi-DTW's matching function may help mitigate false alignments from symmetric or reversed scenarios. Targeted data augmentation and sampling techniques could also help differentiate such challenging cases.

\vspace{-0.3cm}

\vspace{-0.3cm}

\subsection{Ablation study: Effectiveness of triplet loss}

In this ablation study, shown as Figure \ref{table-ablation-2}, we evaluated the performance of different variations of the proposed model, M3, on the CC\_WEB\_VIDEO dataset. The results, as shown in the table, indicate that the incorporation of the triplet loss significantly improves the model's performance, achieving 54.1\% mAP by class and 65.7\% mAP by clip. Moreover, when pretraining is added to the model, the performance is further improved, achieving 84.5\% mAP by class and 89.2\% mAP by clip. These results suggest that the triplet loss is crucial for a similar video retrieval task, and pretraining enhances the model's generalization ability.

\begin{table}[htbp]
\caption{effectiveness of triplet loss.}
\label{table-ablation-2}
\begin{center}
\begin{scriptsize}
\begin{sc}
\begin{tabular}{lcc}
\toprule
Model  & \multicolumn{2}{c}{CC\_WEB\_VIDEO}    \\
       & by class  & by clip  \\
\midrule
M3 autoencoder                             & 17.6\%  & 32.2\% \\
M3 autoencoder + triplet loss              & 54.1\%  & 65.7\% \\
M3 autoencoder + pre-train                 & 29.6\%  & 35.1\% \\
M3 autoencoder + triplet loss + pre-train  & \textbf{84.5\%}  & \textbf{89.2\%} \\
\bottomrule
\end{tabular}
\end{sc}
\end{scriptsize}
\end{center}
\end{table}

\subsection{Ablation study: Effectiveness of bi-directional DTW}

In this ablation study, we evaluated the performance of the proposed model, M3, with two different dynamic time warping (DTW) methods: vanilla DTW and bi-directional DTW (Bi-DTW). The results presented in the table show that the overall performance of M3 improves significantly when utilizing the Bi-DTW method. Specifically, the model trained with Bi-DTW achieves 84.5\% accuracy by class and 89.2\% accuracy by clip level, whereas the model trained with vanilla DTW only achieves 72.1\% accuracy by class and 85.4\% accuracy by clip. This suggests that utilizing the Bi-DTW method dramatically enhances the model's performance for similar video retrieval.

\begin{table}[htbp]
\caption{effectiveness of bi-directional DTW algorithm.}
\label{table-ablation-3}
\begin{center}
\begin{scriptsize}
\begin{sc}
\begin{tabular}{lcc}
\toprule
Model  & \multicolumn{2}{c}{CC\_WEB\_VIDEO}    \\
       & by class  & by clip  \\
\midrule
M3 + vanilla DTW     & 72.1\%  & 85.4\% \\
M3 + Bi-DTW  & \textbf{84.5\%}  & \textbf{89.2\%} \\
\bottomrule
\end{tabular}
\end{sc}
\end{scriptsize}
\end{center}
\end{table}

Reflecting on the findings of our ablation studies, it becomes evident that the choice of methods hinges upon the specific scenario. For tasks necessitating precision and generalization, particularly in complex and intricate video data, it is recommended to incorporate pretraining and challenging sample retraining. These techniques significantly boost performance but require substantial computational resources, indicating their appropriateness in scenarios where such resources are available and accuracy is paramount. Meanwhile, the use of triplet loss has been proven to be effective across tasks, delivering substantial performance improvements. However, for tasks where even more enhanced results are sought, it is beneficial to couple the triplet loss technique with pretraining. When dealing with dynamic time warping, Bi-DTW stands out as a superior choice over the vanilla DTW, especially when striving for accuracy in similar video retrieval, making it the go-to method in such contexts.

\section{Conclusion}

In this study, we introduced a bi-directional dynamic time-warping technique for video information retrieval and innovative strategies for 3D video inquiries, including a novel 3D network architecture and a method to manage 3D video data with added depth. Our models, especially M3, demonstrated superior performance on both 2D and 3D datasets, outpacing both non-deep and deep learning baselines. When evaluating performance, it's essential to consider the aggregation level and the nature of the dataset. Moreover, incorporating pretraining, challenging second round retraining, and our proposed time-warping technique is crucial for optimal performance. Our methods not only offer advanced solutions for video retrieval but also pave the way for improved 3D video data processing, presenting a promising direction for future research.

\bibliography{references_updated}{}
\bibliographystyle{IEEEtran}

\end{document}